\documentclass[letterpaper, 10 pt, conference]{IEEEtran}
\IEEEoverridecommandlockouts
\UseRawInputEncoding

\usepackage{balance}
\usepackage{cite}
\usepackage{amsmath,amssymb,amsfonts,bbm,mathtools}
\usepackage{hyperref}
\usepackage{graphicx,subcaption}
\usepackage{textcomp}
\usepackage{xcolor}
\usepackage{comment}
\usepackage{multirow}

\usepackage{amsmath, amssymb , graphicx, subcaption}
\usepackage{multirow}
\usepackage{dblfloatfix} 
\usepackage{balance}
\usepackage{hyperref}
\usepackage{enumitem}

\DeclareCaptionFont{caption}{\fontsize{8}{9.6}\selectfont}
\usepackage[showframe=false,left=0.75in, right=0.75in, top=1in, bottom=0.75in]{geometry}

\title{\LARGE \bf
CRANE: a 10 Degree-of-Freedom, Tele-surgical System for Dexterous Manipulation within Imaging Bores}

\author{Dimitri Schreiber$^1$, Zhaowei Yu$^1$, Hanpeng Jiang$^1$, Taylor Henderson$^{1,\dag}$, Guosong Li$^{1,\dag}$, Julie Yu$^{2,\dag}$,\\ Renjie Zhu$^{1,\dag}$, Alexander M. Norbash$^3$, and Michael C. Yip$^1$, \IEEEmembership{Senior Member, IEEE}
\thanks{
\dag equal contribution,
$^1$Department of Electrical and Computer Engineering, University of California San Diego, La Jolla, CA 92093 USA. {\tt\small \{dschreib, zhy125, hjiang, g4li, rezhu, tjwest, m1yip\}@eng.ucsd.edu}
$^2$Department of Mechanical Engineering, University of California San Diego, La Jolla, CA 92093 USA. {\tt\small jhy015@ucsd.edu}
$^3$Department of Radiology, University of California San Diego, La Jolla, CA 92093 USA. {\tt\small anorbash@ucsd.edu}}}

\begin{document}

\maketitle
\thispagestyle{empty}
\pagestyle{empty}

\begin{abstract}
Physicians perform minimally invasive percutaneous procedures under Computed Tomography (CT) image guidance both for the diagnosis and treatment of numerous diseases. For these procedures performed within Computed Tomography Scanners, robots can enable physicians to more accurately target sub-dermal lesions while increasing safety. However, existing robots for this application have limited dexterity, workspace, or accuracy. This paper describes the design, manufacture, and performance of a highly dexterous, low-profile, 8+2 Degree-of-Freedom (DoF) robotic arm for CT guided percutaneous needle biopsy. In this article, we propose CRANE: CT Robot and Needle Emplacer. The design focuses on system dexterity with high accuracy: extending physicians' ability to manipulate and insert needles within the scanner bore while providing the high accuracy possible with a robot. We also propose and validate a system architecture and control scheme for low profile and highly accurate image-guided robotics, that meets the clinical requirements for target accuracy during an in-situ evaluation. The accuracy is additionally evaluated through a trajectory tracking evaluation resulting in $<$0.2mm and $<$0.71$^{\circ}$ tracking error. Finally, we present a novel needle driving and grasping mechanism with controlling electronics that provides simple manufacturing, sterilization, and adaptability to accommodate different sizes and types of needles.



\end{abstract}

\section{INTRODUCTION}

Within the field of Image Guided Surgery, Intraoperative CT guidance is used to guide the physician to both diagnose and treat numerous diseases, primarily where there are complex 3D anatomical constraints and small, deep, target nodules that reside typically 10 cm or more below the surface of the ski~ \cite{Tsai2009, ctguided_retrospective}. Three of the procedures most frequently treated via this approach are biopsies and ablations of lung, liver, and kidney tumors: all in the abdominal area. 

\begin{figure}[h]
    \centering
      \includegraphics[width=\linewidth]{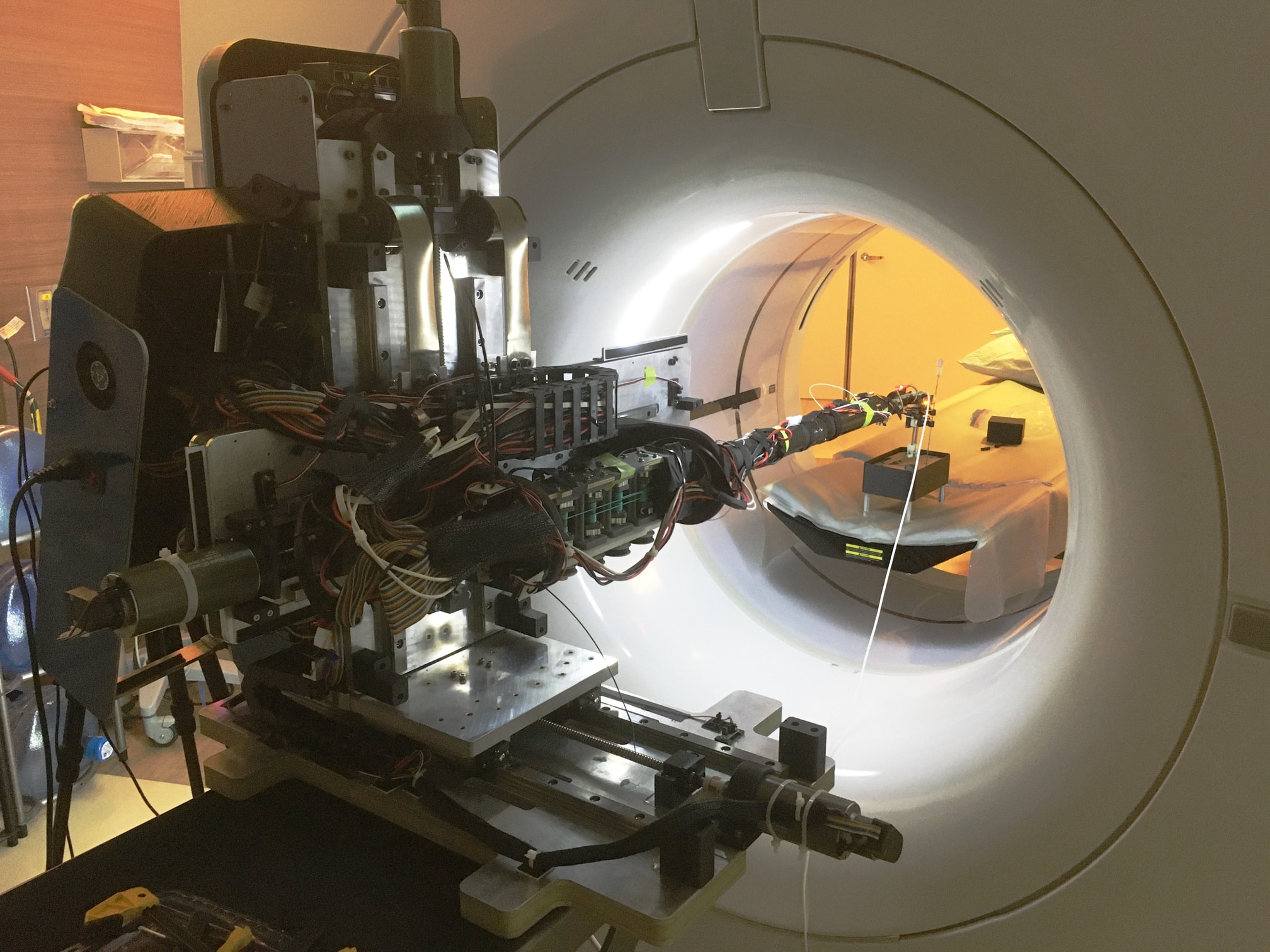}
    \caption{CRANE is a CT compatible biopsy robot featuring high dexterity, high accuracy and low in-bore profile. These features allow physicians to perform their procedures with improved efficiency and safety.}
    \label{fig:robot_person_UCSD}
    \vspace{-5mm}
\end{figure}
During an intraoperative CT guided procedure, the physician must localize and successfully insert a needle-like probe into a target. Typically, the physician alternates between advancing the needle a fractional distance to the target, stepping away, and scanning the patient to receive an update on the needle and tumor's position. When advancing the needle, the patient is withdrawn from the scanner bore to allow the physician to ergonomically place the needle. The combination of the non-real-time nature of CT scans, the lack of tool tracking, and the freehand adjustments result in the physician needing to cognitively visualize the 3D trajectory and estimate the amount of fine needle adjustment required to hit the target, all without visual feedback. This results in an increased number of scans needed to hit the target precisely, greater variability in and lengthier procedure times, and potentially need to withdraw and re-puncture the chest wall. These factors result in decreased safety with clinically relevant side effects \cite{heerink2017complication, Accordino2015}, decreased positive predictive value \cite{wang2016ct} and increased radiation dose \cite{national2013results}. Furthermore, small, deep lesions are especially difficult to reach as slight angulation errors result in significant tip positioning errors\cite{Tian2017}. Robotics offers a potential solution to these issues by providing increased accuracy for needle trajectories and allow removing the tedious back and forth procedure flow, providing the opportunity for physicians to precisely insert the percutaneous devices along more cranial-caudal trajectories rather than the in-plane trajectories currently selected \cite{kettenbach2015robotic}. Existing robotic platforms have limitations in dexterity, work-space, accuracy, device-setup, or instrument compatibility.

This paper introduces the CT Robot and Needle Emplacer, CRANE (Fig. \ref{fig:robot_person_UCSD}), a needle-manipulation system that is designed to focus on dexterity, workspace, and accuracy within an imaging bore of a scanner. CRANE’s novel design 
demonstrates a fully active, serially linked, redundant kinematic approach with closed-loop tip pose control with a new needle-grasper which allows for large dexterity within an imaging bore with high accuracy, while being able to accommodate insertion and retraction of a variety of needle-like surgical tools.

We provide the following technical contributions:
\begin{enumerate}
    \item Low profile, redundant, and dynamic serial link design with high dexterity and actuator bandwidth.
    \item High tip-accuracy achieved through multi-level control to compensate for errors in individual system components. Analysis of errors is presented.
    \item Development of a novel disposable and sterilizable clutching needle driver using SMA actuator providing a simple needle grasping mechanism.
\end{enumerate}

\section{Related Work}
Needle insertion robotic platforms for use within imaging bores have been developed for a wide range of applications across the human body, ranging from leg-bone biopsy to brain surgery. Of these applications and platforms, this overview focuses on systems applied to the torso and chest region, frequently the most restrictive of anatomy with regards to in-bore space as patient chest and abdomens are the largest area of the anatomy. Additionally, procedures performed within this region have a large number of anatomical obstacles, including the rib cage and large blood vessels, and face significant anatomical motion due to lung motion, large blood vessel pulsating, and the digestive system. 

Needle insertion platforms for use around patient torsos within imaging bores can be broadly grouped into two clusters based on if they are mounted to the patient or mounted to the floor and scanner bed. Within each of these categories, systems can be fully active or rely on passive setup and joints. 

Patient mounted systems are typically smaller and under-actuated \cite{Walsh2008, Ghelfi2018, Hiraki2017, Maurin2006, Yang2017, Wu2019}, however, a select few are fully actuated \cite{Bricault2008, Hungr2016, xact_robotics}. These systems naturally move with the patient, which provides benefits through inherent system motion with patient respiration. Bore mounted systems \cite{Stoianovici1998, bio-rob2, Schulz2013, Stoianovici1997, Masamune2001} with passive setup joints or a mixture of passive and active joints provide a compromise between the challenges of patient setup while maintaining a fairly low bore profile and retaining high system stiffness. However, they have limited ability to compensate for gross patient motion due to pain or coughing and target motion within the body due to physiological motion. Additionally, for both patient mounted systems and those utilizing passive setup joints, the manual positioning and attachment to the patient or manual understanding and positioning of the setup joint can be challenging, add time and complexity to a procedure, and preclude certain superior needle insertion trajectories if a setup pose can not be found.

Fully active systems (either floor \cite{Yang2010a, Hiraki2017, Tovar-Arriaga2011, jhu_fully_active_biorob, Fichtinger2002, korea_cool_robot} or table mounted \cite{Moreira2017a, Shahriari2015, Stoianovici2003, number25, frishman2021, Bricault2008}) provide numerous advantages, both in resolving the aforementioned issues (decreased setup complexity, ability to regulate tip stiffness decreasing tissue damage due to respiratory motion) with patient-mounted systems and those utilizing passive setup joints and beyond, including the ability to minimize physical contact with the patient decreasing a rising concern with the COVID-19 pandemic. Systems using large industrial arms with a custom end-effector \cite{Tovar-Arriaga2011} provide high stiffness at the cost of a larger system size with less intra-bore dexterity and workspace available and limited ability to manually remove the system in case of a system failure. These systems operate outside the bore rather than within due to space considerations. 

These aforementioned systems use a variety of methods of interacting with the needle, ranging from passive needle guides \cite{Stoianovici1998, bio-rob2, jhu_fully_active_biorob, Schulz2013, Stoianovici1998, Masamune2001, Fichtinger2002, Stoianovici2003, number25, Bricault2008} to a variety of active mechanisms including fixed travel insertion \cite{Yang2010a, kapoor2008, Schreiber2019}, rollers \cite{solomon2002robotically}, clutches \cite{frishman2021, Frishman2020}, graspers \cite{Ghelfi2018}, and one-time use gear-wrapper guides \cite{xact_robotics}. These designs have limitations in their compatibility with needles and probes, high complexity, potentially challenging sterilization procedures, and have demonstrated damage to fragile ablation probes \cite{solomon2002robotically}. 
A fully active and low-profile system presented in \cite{Schreiber2019}
demonstrated large workspace, high dexterity, and high physician-in-loop
accuracy. This design focused on moving actuators outside of the bore via a cable-drive transmission coupled into the scanner via a thin carbon-fiber tube. However, due to the long kinematic chains, long travels of cable driven joints with numerous pulleys, and lack of joint and tip sensing, robot state estimation and system accuracy are low. Additionally, due to non-optimal cable routing, the joint ranges are limited and the maximum active needle insertion depth is only 50mm. 

CRANE supersedes this design with a fully re-engineered system which greatly increases the work-space via the addition of a vertical axis, optimized cable routing providing over 40$\%$ higher cable-driven joint travels, and unlimited needle insertion length via a clutching needle grasper. These previous travel and insertion length values were found to be limiting upon physician-user evaluation. Furthermore, this new system provides high end-effector accuracy through the use of multi-loop controllers and multi-level joint and end-effector sensors and provides improved safety.

\section{Design Requirements and Specifications}
CT-compatible biopsy robots need to follow certain clinical requirements, guiding the design of these systems:

\begin{itemize}[leftmargin=1em]
    \item \textit{Forces}: a maximum of 10N needle insertion force with 0.06Nm torques to adjust needle orientation while moving through tissue \cite{Poniatowski2016, Walsch_thesis_2010, Walsh2008}. This requirement is achieved through the use of a rigid base structure, high strength plastic in-bore joints with high strength synthetic cables and a strong needle grasping mechanism.
    \item \textit{Workspace}: the system should be able to insert needles across the human body in different configurations without colliding with the patient's body or the scanner bore. 
    This requirement is achieved by having low profile in-bore components for the system with large travels outside the bore. 
    This design significantly improves robot dexterity over previous robotic systems \cite{Schreiber2019}.
    
    \item \textit{Precision}: Abdominal needle insertion procedures typically require $<2mm$ position and $<2^{\circ}$ angular accuracy. Here, this requirement is achieved through a low-backlash transmission with joint level encoders, closed loop end-effector control, and validated via trajectory tracking tests.
    \item \textit{Needle Interface}: Physicians use a variety of needle-like probes during their practice, and these needles should be quick to attache and remove. 
    This requirement is achieved through a novel, disposable, removable, and sterilizable mechanical needle interface using helically wrapped Shape Memory Alloy actuators
    \item \textit{Image Artifacts}: the system should not cause major imaging artifacts. This is achieved through the use of plastics, composites, and ceramics in the bore with minimal use of high-density materials within the scanning area.
\end{itemize}


\begin{figure}[t]
    \centering
      \includegraphics[width=\linewidth]{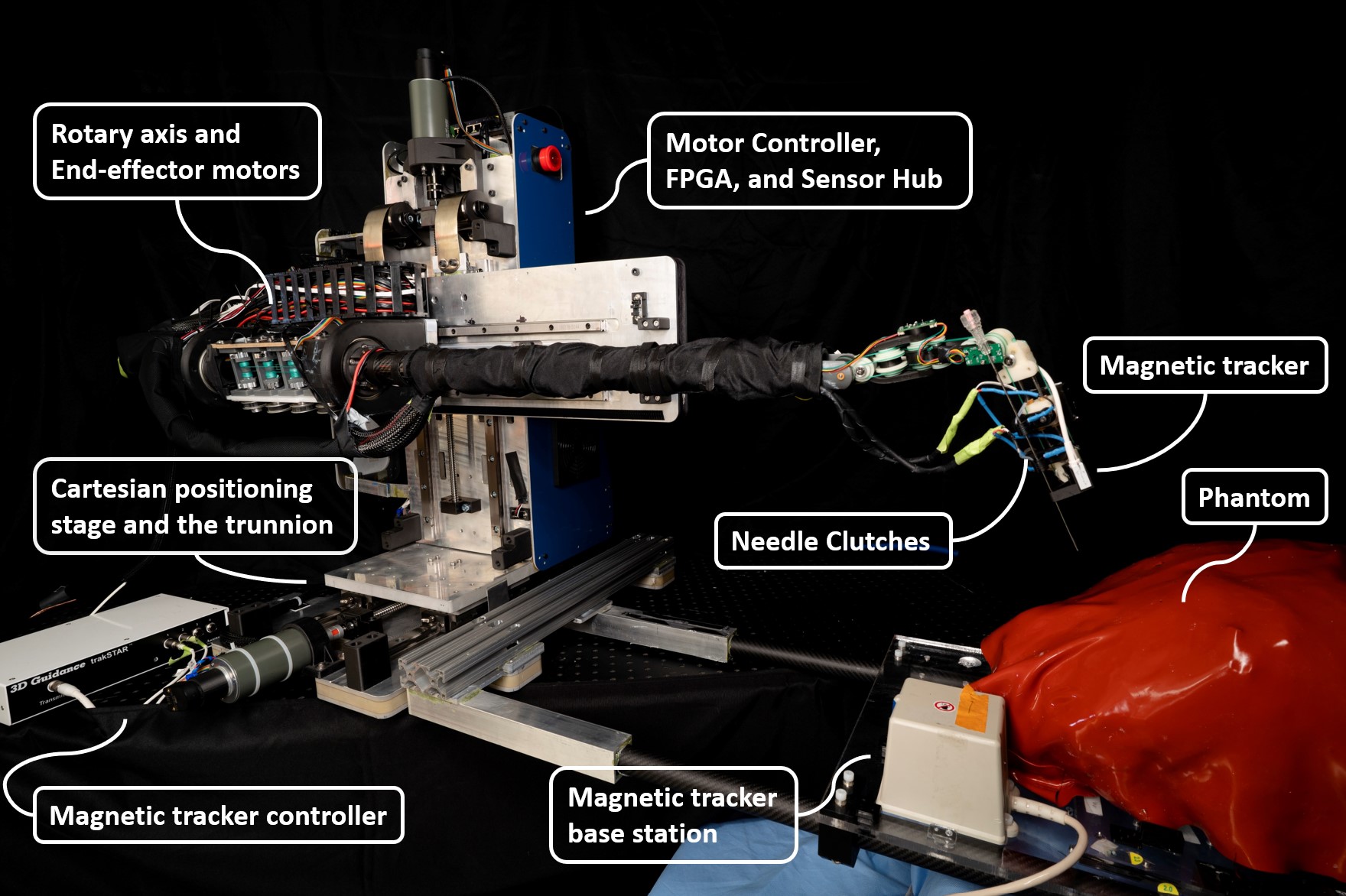}
    \caption{Overview of the robot platform experimental setup including CRANE system, magnetic tracker, and needle-biopsy phantom. The joints are remotely driven via cables running from motors housed on the base motion stage to the end-effector through the carbon fiber tube. External connections to the system are limited to a three cables: an Ethernet connection, a USB connection, and AC power input.}
    \label{fig:robot_desc}
    \vspace{-5mm}
\end{figure}

\section{Mechanical Design}

%

\subsection{In-Bore Mechanical Design}
The philosophy guiding this system's mechanical design, illustrated in Fig. \ref{fig:robot_desc}, is to minimize in-bore volume to maximize space for needle insertion and the patient within the scanner. In-bore joints account for 4-DoF utilize remote actuators with a 2N cable drive to maintain a low weight, inertia, and volume. The joints are manufactured with Carbon Fiber Reinforced Plastic, ZrO$_2$ ceramic bearings, bushings, 1.25mm diameter Dyneema SK99 synthetic cables, and minimal metallic components to prevent imaging artifacts. 
Through optimized cable routing, including decreased pulley-to-pulley clearances and improved pulley positioning, the cable-driven revolute joint travels reach 200 degrees and are limited by self-collision with neighboring links. 
Capstan drive pulleys are connected to a 44:1 geared Maxon motors via GT2 timing belts with a final drive reduction of 219.7:1.
The capstan to joint drive ratio reduction is 2.27:1, providing a final drive line reduction of 219.7:1 and resulting in a maximum independent joint velocity of $5.16\text{rad}/s$. Such velocity greatly exceeding the requirement for a needle manipulation robot to compensate for anatomical motion; the $3.36\text{N/m}$ joint torque provides sufficient end-effector force in all configurations.

The needle linear-insertion joint's length is short, allowing operation in tight space between the scanner bore and patient who has large body habitus.
Although the insertion joint's length is shorter, the maximum needle insertion length is increased from the addition of a novel needle insertion clutching mechanism. 

FEA analysis of the in bore structure excluding cables of the robot was performed with a needle insertion force applied to the end-effector resulting in a 10.3mm suggested tip deflection. This deflection motivates the direct end-effector sensing and feedback control.

\begin{figure}[b!]
    \centering
    \vspace{-2mm}
      \includegraphics[width=\linewidth,clip=true,trim={0 0 0 15mm}]{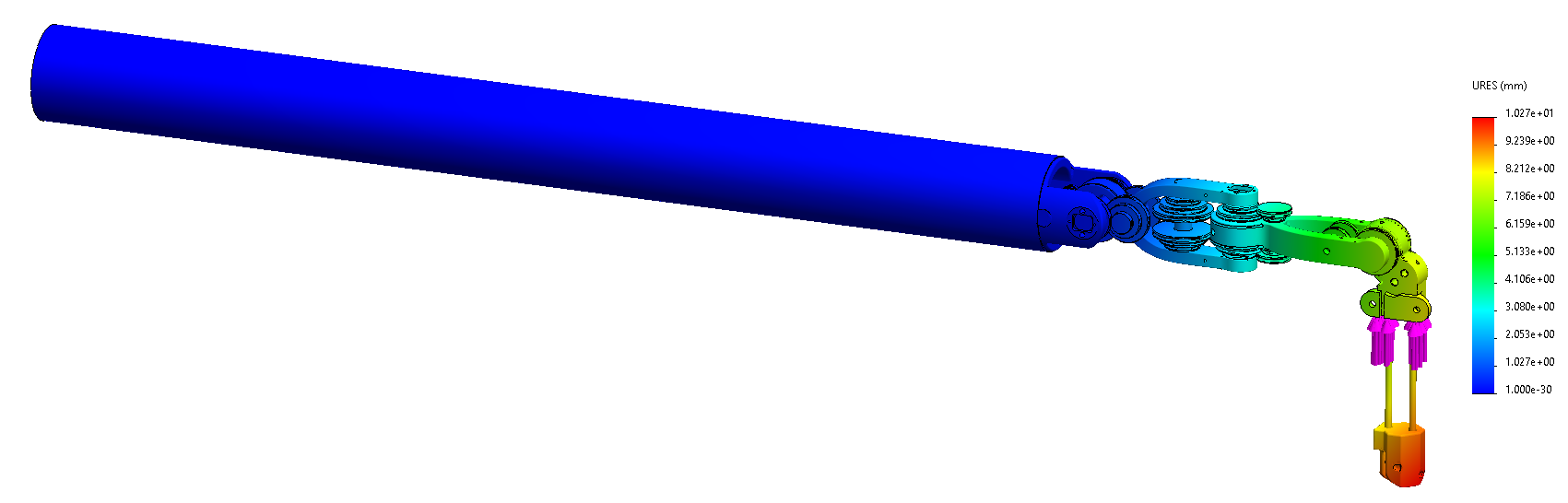}
    \caption{FEA of robot end-effector neglecting cable elasticity to illustrate structural deflection which joint-level controllers are unable to compensate for. Simulated with a 10N force applied to the end-effector demonstrating 10.3mm of deflection at the tip.}
    \label{fig:FEA}
\end{figure}

\subsection{Base Positioning Stage Mechanical Design}
The base motion stage provides 4-DoF and houses the motors for the in-bore joints. The base motion stage consists of a high stiffness base 3-DoF cartesian motion stage and a 1-DoF rotary trunnion, labeled in Fig. \ref{fig:robot_desc}. 
Gravity counterbalance on the vertical axis is achieved through constant force springs (McMaster 9293K69). 
Maximum linear velocities of 0.166 m/s, 0.166 m/s, and 0.332 m/s are achievable for the first three joints. Joint ranges are 400mm for all axis.

Hardware stiffness and backlash evaluation were performed using a dial indicator (Shars, 0.0005") and Z-style load cell (500N range) with stiffness measured as $>100\text{N/mm}$ and backlash $<0.1\text{mm}$. These values far out-perform clinical requirements for target procedures. Therefore, the base motion stage is neglected from the system stiffness analysis.





\subsection{Needle Clutch}
\begin{figure}[t!]
    \centering
      \includegraphics[width=\linewidth]{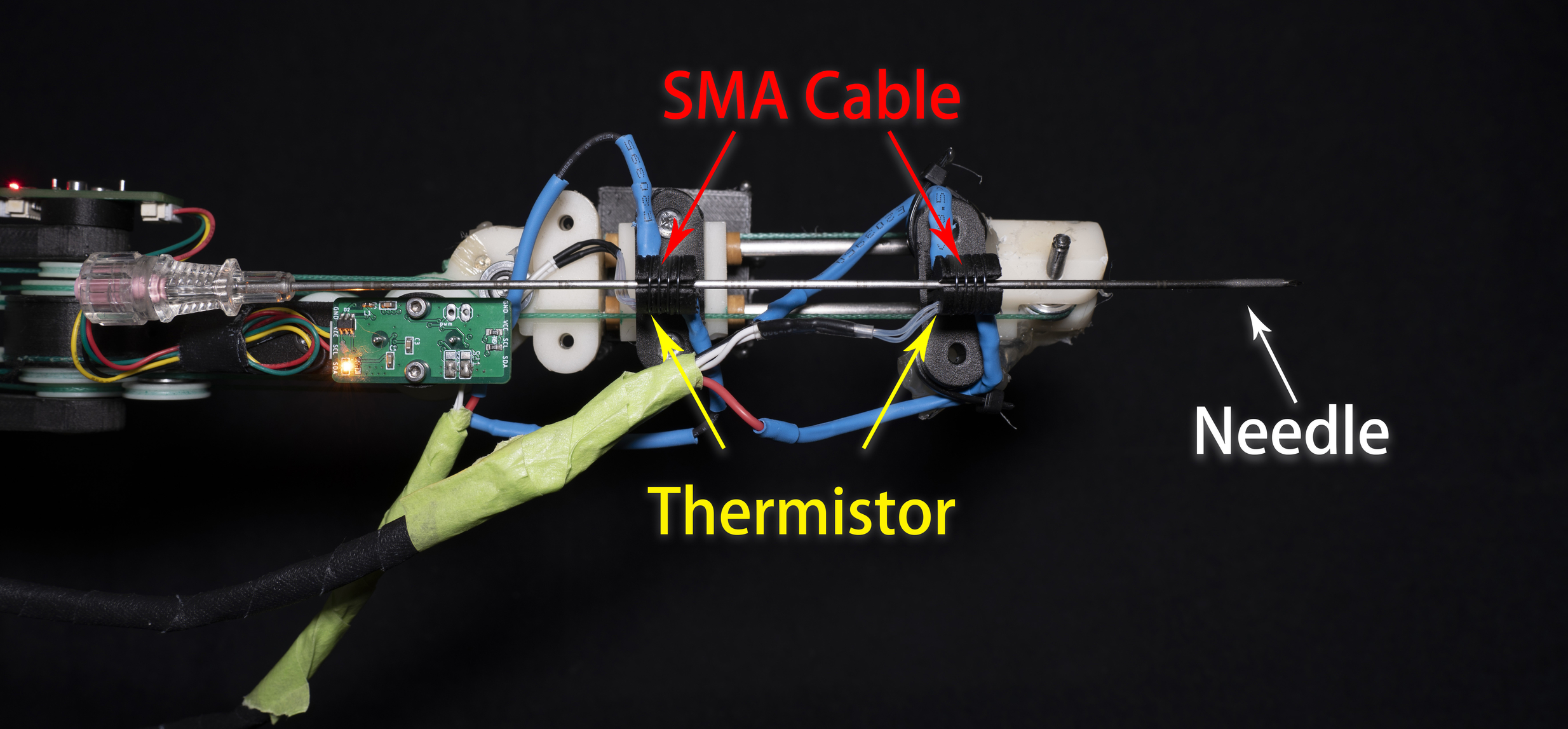}
    \caption{Illustration of the clutch mechanism with labeled components. The flexure is 3D printed in carbon fiber reinforced nylon (Markforged Onyx) with a heat-deflection (ASTM D648 B) temperature of $145^oC$. A thermistor is bonded directly underneath the clutch for temperature measurement, closed-loop control, and safety.}
    \label{fig:clutch_components}
    \vspace{-4mm}
\end{figure}

Physicians frequently need to insert needles deep into the body in order to reach anatomical targets. However, a full length linear needle insertion axis may be undesirable due to its large size and potential for collision with large body habitus patients. Our clutch design, illustrated in Fig. \ref{fig:clutch_components} uses Shape Memory Alloy (SMA) wire actuators (Flexinol 0.015in) wrapped around a plastic flexure clamp to tightly grasp a needle. Through alternating between tightening and loosening the two clutches, the needle could move in an "inch-worm" fashion during an insertion. The clutch could also be deactivated rapidly via active air cooling, allowing the needle to be removed for sterilization or adjusted by hand.
Temperature is measured with a thermistor bonded in place with thermal-epoxy in direct contact with the SMA wire for clutching status.



\section{Electrical and Embedded System}
\begin{figure*}[t!]
    \vspace{-1mm}
    \center
    \includegraphics[width=0.7\textwidth]{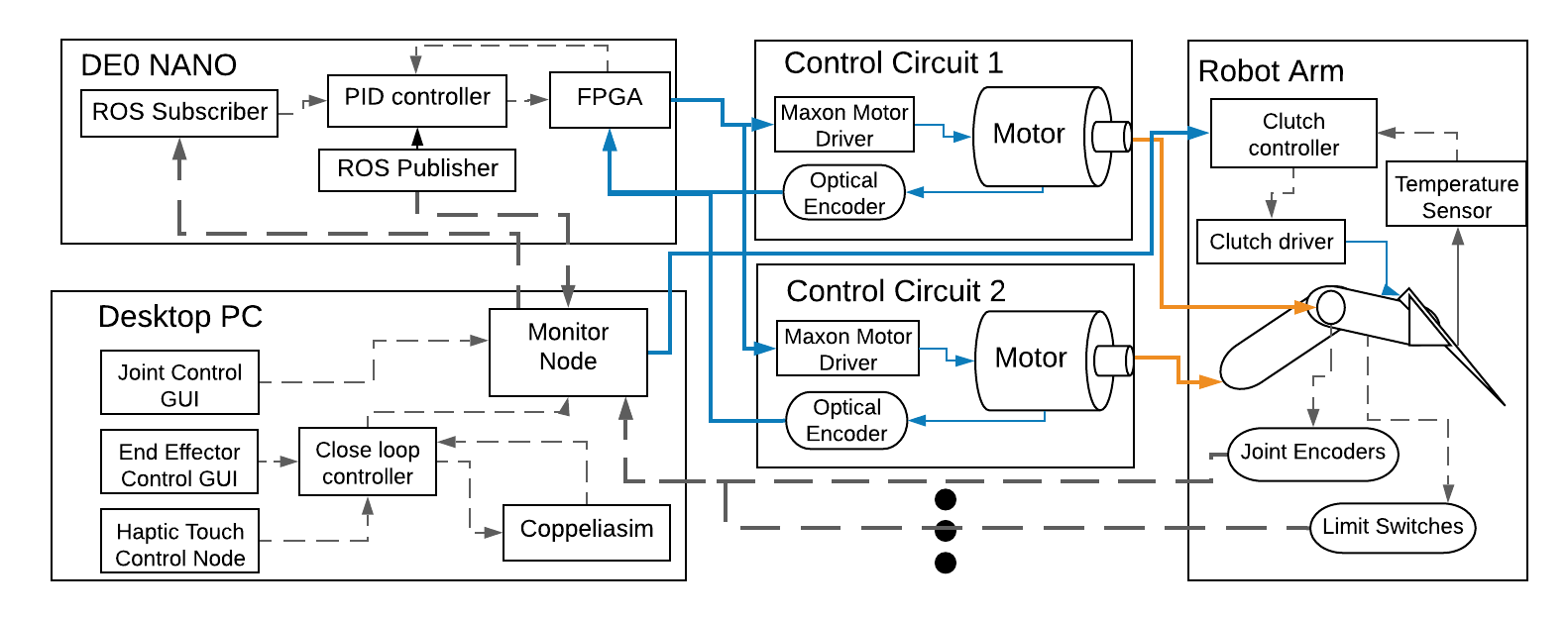}
    \caption{The architecture of the system is divided into a low-level system and high speed controllers (DEO-NANO and control circuits), and a high-level desktop PC component, which contains the higher level low speed controllers and the user interface (Desktop PC). }
    \vspace{-5mm}
    \label{fig:flowchart}
\end{figure*}

The primary electrical circuit of the robots provides power and connections for the DE0-NANO-SoC and Maxon motor drivers, as shown in Fig.\ref{fig:flowchart}. The logic circuit of the board is contained within the Intel FPGA section in the DE0, and it manages the analog signal generation to the motor driver, counting the encoder quadrature signal and a watchdog timer that disables the motors if the update time exceeds $10m s$. The limit switches and joint encoders are being interfaced using a Cypress PSoC and a Cortex-M7 MCUs. These two MCUs pass the sensor values back to the desktop computer via UDP using Ethernet. The Linux section on the DEO has a real-time kernel and runs a 1 kHz PID position with the FPGA section via shared memory and communicated with the desktop PC over ROS with a round trip delay of less than $500 \mu s$.

The SMA needle clutch is heated via Joule heating, and a temperature controller is designed to maintain its target activation temperature using a PID loop on each clutch. 
The circuit utilizes TVS and Zener diodes, fuses, sense resistors, and ferrite-bead based filters to insure correct current control under all conditions.

\section{High Level Controls}
\subsection{Software Architecture and User Interface}
The master control software running on the high-level desktop is a ROS Qt5 based graphical user interface called the monitor node that filters and offset inputs from other nodes contain other user input options and publishes out the finalized joint setpoints. The other user input options include a joint control mode, end-effector control mode, and Touch Haptic Control mode, each running in a separate node and only one could be active at a given time. The joint control node GUI allows the user direct joint control via on-screen buttons and visual monitoring through plots of each joint motor; both the end effector (EE) control node GUI and Touch Haptic Control node allows the user to set a desired EE position and orientation via button press or hand motion. The desired setpoint would be sent into the simulation software, in which sums up position setpoint and target quaternion and to calculate the kinematics needed to control the EE position.

\subsection{Kinematics}
The robot's kinematics chain is described using the Modified Denavit-Hartenberg (DH) parameters, shown in Table \ref{tabl:dh_parameters}. The DH parameters define the position of the next frame relative to the previous frame
\vspace{-1mm}
\begin{equation}
\vspace{-1mm}
    c_{n+1} = c_n + a x_n + d z_{n+1}
\end{equation}
and orientation relative to the previous frame
\begin{equation}
    R_{n+1} = 
    \setlength{\arraycolsep}{3pt}
    \renewcommand{\arraystretch}{0.8}
R_n \begin{bmatrix}
1  & 0  & 0 \\
0  & \cos \alpha  & -\sin\alpha \\
0 & \sin \alpha & \cos \alpha  \\
\end{bmatrix}
\begin{bmatrix}
\cos \theta   & -\sin \theta  & 0 \\
\sin \theta   & \cos \theta  & 0\\
0 & 0 & 1  \\
\end{bmatrix}
\end{equation}
where $R_n$ is the orientation and $c_n$ is the position of the $n$th frame relative to the robot's base frame, $B$ and $q_n$ is the position of the $n$th joint. For this robot with 8 revolute and prismatic axes, $n\in \{1,...,9\}$ with $n = 9$ as the robot's end-effector frame and $n = 1$ as the robot's base frame.  The robot base to tip, also referred to as end-effector (EE), transform is defined as
\begin{equation}
    ^{\text{B}}T_{\text{Tp}}^{\text{B}} = fk(q) \quad \text{for} \quad q \in \mathbb{R}^8
\end{equation} 
define the joint positions, $fk(q)$ is defined by chaining together the link transforms described by the DH convention, B is the robot base coordinate frame and the default coordinate frame for transforms if unlisted, and $^AT_B^C$ describes a $4\times 4$ homogenous transform $\in$ SE(3) from coordinate from $A$ to coordinate frame $B$ relative to base frame $C$.

\begin{table}[b]
    \setlength\tabcolsep{0.9em}
    \centering
    \vspace{-2mm}
    \caption{Modified DH parameters for CRANE where p is a prismatic joint and r is a revolute joint}
    \scriptsize
    \begin{tabular}{c|c|cccc}
          Frame & Type & a (meters)& $\alpha$ (rad) & D (meters) & $\theta$ (rad) \\ \hline
          1 & p & 0 & $- \frac{\pi}{2}$ & $q_1$ & $0$ \\
          2 & p & 0 & $- \frac{\pi}{2}$ & $q_2$ & $- \frac{\pi}{2}$ \\
          3 & p & 0 & $- \frac{\pi}{2}$ & $q_3$ & $- \frac{\pi}{2}$ \\
          4 & r & 0 & 0 & 0 & $q_4$ \\
          5 & r & 0 & $\frac{\pi}{2}$ & 0 & $q_5 + \frac{\pi}{2}$ \\
          6 & r & 7e-2 & $\frac{\pi}{2}$ & 0 & $q_6$ \\
          7 & r & 7e-2 & $\frac{\pi}{2}$ & 3e-2 & $q_7 - \frac{\pi}{2}$ \\
          8 & p & 1e-2 & $-\frac{\pi}{2}$ & $q_8$ & 0 \\
          9 & - & 0 & 0 & 6e-2 & $\frac{\pi}{2}$ \\

    \end{tabular}
    \label{tabl:dh_parameters}
\end{table}

The motor and joint positions, $\theta, q \in \mathbb{R}^8$, are related as
\begin{equation}
    q=L\theta
\end{equation} where $L$ is the $8\times8$ coupling matrix. In the case of joint mounted actuators or an uncoupled transmission, $L$ is diagonal and corresponds to the simple gear-ratio of the transmission, as in our $q\in\{1,...,4\}$. In coupled situations, $L$ is upper triangular. Due to manufacturing tolerances, $L$ is constructed during a calibration step from data row-wise and calculated as a least-squares linear-regression problem as 
\begin{equation}
L_{i,*} = q_{i,*} \theta^{\dagger} \quad \text{where}\quad \theta^{\dagger}=\theta^{T}\left(\theta\theta^{T}\right)^{-1}
\end{equation}
 for each row $i$ of $L$ with $q_{i,*}$ being a time series of $m$ samples a single joint's angle and $\theta \in \mathbb{R}^{8 \times m}$ being a time series of all motor angles being used as inputs for the coupling matrix for $8$ output joints.  Here, joint $q \in \{1,...,4\}$ are calibrated individually as scalar terms and $q \in \{ 5,..,8 \}$ are calibrated together as a matrix.This matrix can be calculated analytically from the system design or empirically off observed data, but by doing it empirically, errors between ideal and actual kinematic parameters are reduced. 

\subsection{End-Effector and Joint Control}

The estimated joint state, $q_{\text{est}}$, is constructed via a complementary filter between the motor's velocity, $\dot{\theta}$, and the magnetic joint encoders position, $q_{\text{meas}}$, as
\begin{equation}
    q_{\text{est}}=\alpha L \dot{\theta}_{\text{meas}} \Delta T + (1- \alpha ) q_{\text{meas}}
\end{equation} for a sampling time, $\Delta T$,  and weighting parameter, $\alpha$, corresponding to the changeover frequency of the filter between the two sensors. The complementary filter helps to reduce errors resulting from high frequency noise in magnetic joint encoder readings and the coupling matrix equation's errors due to the cable-transmission's spring-stiffness.

This joint angle estimate, $q_{\text{est}}$ is used to update the motor set-point position, $\theta_{\text{set}}$ following a PD control law in the joint space 
\begin{equation}
        \theta_{\text{set}} \leftarrow \theta_{\text{set}} + \Delta \theta \quad 
        \text{for} \quad
        \Delta \theta = L^{-1} \left(K_p e_q + K_d \frac{d e_q}{ d t } \right)
\end{equation}
where $\quad e_{q}=q_{\text{set}}-q_{\text{est}}$, $q_{\text{set}}$ is the joint angle setpoint, and $K_p,K_d$ are the proportional and derivative gains. 
EE pose errors are calculated for position and orientation as

\begin{equation}
\begin{multlined}    
    \qquad e_{\text{pos}}  = x_{\text{targ}} - x_{\text{meas}} \text{, and} \\
    e_{\text{ori}} = \angle(z_{\text{targ}}    z_{\text{meas}})(z_{\text{targ}} \times z_{\text{meas}}) \qquad
\end{multlined}
\end{equation}
where
\vspace{-3mm}

\begin{equation}
\vspace{-3mm}
\angle(z_{\text{targ}}, z_{\text{meas}}) = \cos^{-1} \left( \frac{z_{\text{targ}}^T z_{\text{meas}}}{\|z_{\text{targ}}\|_2 \|z_{\text{meas}}\|_2} \right).
\vspace{3mm}
\end{equation}
\noindent $x_{\text{targ}}$ and $x_{\text{meas}}$ are target and measured 
translation vectors of the target tip transform, $^{\text{B}}T_{\text{Tpt}}$, and measured tip transform,
$^{\text{B}}T_{\text{Tpm}}^{\text{}}$. $z_{\text{targ}}$ and  $z_{\text{meas}}$ are the
Z axis vectors of the rotation sub-matrix of
$^{\text{B}}T_{\text{Tpt}}$ and $^{\text{B}}T_{\text{Tpm}}^{\text{}}$. 
The target transform described by $^{\text{B}}T_{\text{Tpt}}$ is provided from the User Interface. 
The measured tip transform in the robot base frame is calculated as
$^{\text{B}} T_{\text{Tpm}} = {^{\text{B}} T_{\text{Tr}}} 
{^{\text{Tr}} T_{\text{Tp}}}$ where
$^{\text{Tr}} T_{\text{Tp}}$ is the magnetic tip tracker's pose in the tracker's base frame. The transform from the robot's base frame to the tracker's base frame,
$^{\text{B}} T_{\text{Tr}}$, is found by solving a least squares transform  between $^{\text{Tr}}T_{\text{Tpm}}$ and $^{\text{B}}T_{\text{Tp}}$ based on a initialization
sequence. As needles are symmetric, the orientation error does not include rotation around the EE's z-axis.

The joint angle update is calculated as 
\begin{equation}
    \begin{aligned}
    q_{\text{set}} \leftarrow q_{\text{est}} + \Delta q \quad \text{where} \qquad \\
    \Delta q = K_{\text{pos}}J_{\text{pos}}^{\dagger}e_{\text{pos}} + K_{\text{ori}}J_{\text{ori}}^{\dagger} e_{\text{ori}}
    \end{aligned}
\end{equation}
where the current end-effector position and orientation Jacobians, $J_{\text{pos}} \in \mathbb{R}^{3 \times 8}$ and $J_{\text{ori}} \in \mathbb{R}^{3 \times 8}$, come from the forward kinematics and is 
evaluated near the current joint angles $q_{\text{est}}$ and $J^{\dagger}$ is the psuedoinverse calculated via the Levenberg-Marquardt algorithm. $K_{\text{pos}}$ and $K_{\text{ori}}$ are the proportionality control constants for the end-effector controller. 


\section{Experiments and Results}
\subsection{Trajectory Tracking Accuracy Evaluation}
\begin{figure}[t!]
    \centering
      \includegraphics[width=\linewidth,clip=true]{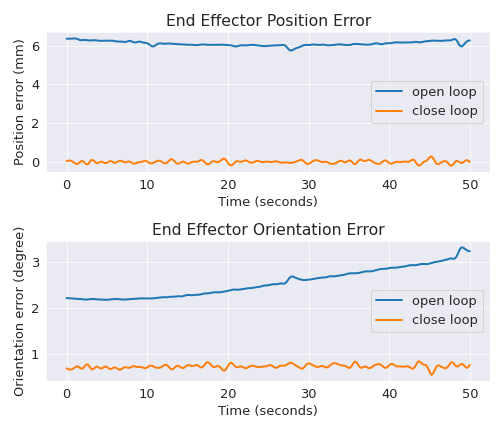}
    \caption{Open-loop evaluation is performed using motor position control without feedback from joint encoders or end-effector controller. Close-loop control runs position control using joint encoders for feedback and direct end-effector position measurement from a magnetic 6DoF position tracker.}
    \label{fig:trajectory_tracking}
    \vspace{-7mm}
\end{figure}

The system's accuracy was evaluated by performing a virtual Remote Center of Motion trajectory where the robot revolved around a virtual needle tip location. Here, the robot's end-effector follows a cone trajectory simulating the workspace a physician would use during an actual procedure. The Ascension TrackStar magnetic tracker was used for accuracy measurement. The mean resulting accuracy, shown as a time series in Fig. \ref{fig:trajectory_tracking}, across the trajectory was $0.27mm$ and $0.71^\circ$.

Two evaluations of the system's trajectory tracking accuracy were performed. In the open-loop test, all joint and end-effector controllers were disabled. Joint angles were purely calculated off the ideal coupling matrix, $K$, without compensation for cable stretch and hysteresis in the in-bore transmission. EE measurements were replaced for the $J^-1$ controller with predicted EE positions based on the forward kinematics of the calculated joint angles from the motor. With controllers disabled, position and orientation errors are significantly increased due to the mixture of joint tracking error, system deflection, and manufacturing errors. Closed loop control using direct end-effector tracking enables the system to accurately reach targets despite these challenges.

\begin{figure}[b!]
\vspace{-7mm}
    \center
    \includegraphics[width=0.45\columnwidth]{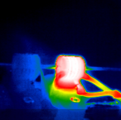}
    \includegraphics[width=0.45\columnwidth]{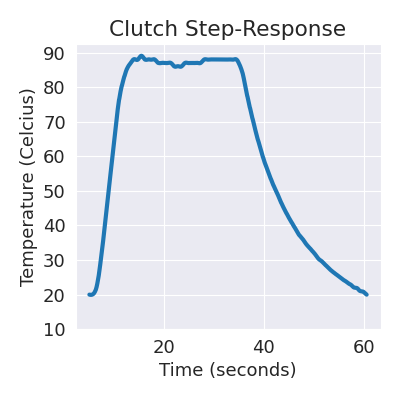}
    \caption{Left: thermal image of the clutching needle driver on the robot end-effector with one clutch activated. Right: a step response collected from the clutch showing Joule heating and air-blast assisted cooling. The highest temperature recorded is $91^{\circ}C$. Clutch is active at $80^o$ C and deactivated at $40^o$ C with a 2.5 second rise time and 10.1 second fall time.}
    \label{fig:clutch_exp}
\vspace{-1mm}
\end{figure}

\subsection{Needle Clutch Evaluation}

The needle clutch is 3D printed in a nylon-carbon composite material on a Markforged 3D printer. Performance with regards to clutching force and cycle time for the clutch was evaluated. The two clutches sized for a 15-gauge needle were activated and deactivated more than 50 times, measuring slipping forces using spring scales (0-50N and 0-5N ranges). Activated slipping forces were measured at 18N and 20N. Deactivated slipping forces were measured at 1N and 2.25N.  

The clutch and driver step-response for on and off motions was evaluated to determine the feasible cycle time during a long travel needle insertion. The results, reported in Fig. \ref{fig:clutch_exp},  show short on and off times.

\subsection{CT Scanner Biopsy Experiment}
The biopsy task was conducted inside a sliding stage CT
scanner (GE Revolution) at the University of California San Diego’s Thornton Hospital. Scan settings of 120 kVp, 300mAs per slice, 0.5-second rotation time were used. Testing was performed on a custom lung phantom, similar to the designs presented in \cite{Scott1992}. The lung phantom consists a plastic resin rib cage with a volume of 12" x 11" x 7" containing a dry preserved pig lung. The remaining space in the rib cage is filled with gel-candle wax to simulate fat and wrapped with two durometers of silicone sheet to simulate muscle and skin. Several different size silicone nodules are planted into the lung to simulate tumors.

An experienced radiologist teleoperated the robot using from the control room (note their silhouettes behind the glass) using our EE control GUI to guide needle to the nodules with CT image data, external camera and Coppeliasim simulation of the robot were used for visual guidance. The needle insertion vector had less than $1.2^\circ$ error from nominal. Interference was not observed between the robot, magnetic tracker, and scanner. Streaking artifacts within the CT image due to the robot's end-effector are minor.

\begin{figure}[t!]
\vspace{1mm}
    \center
    \includegraphics[width=0.49\columnwidth]{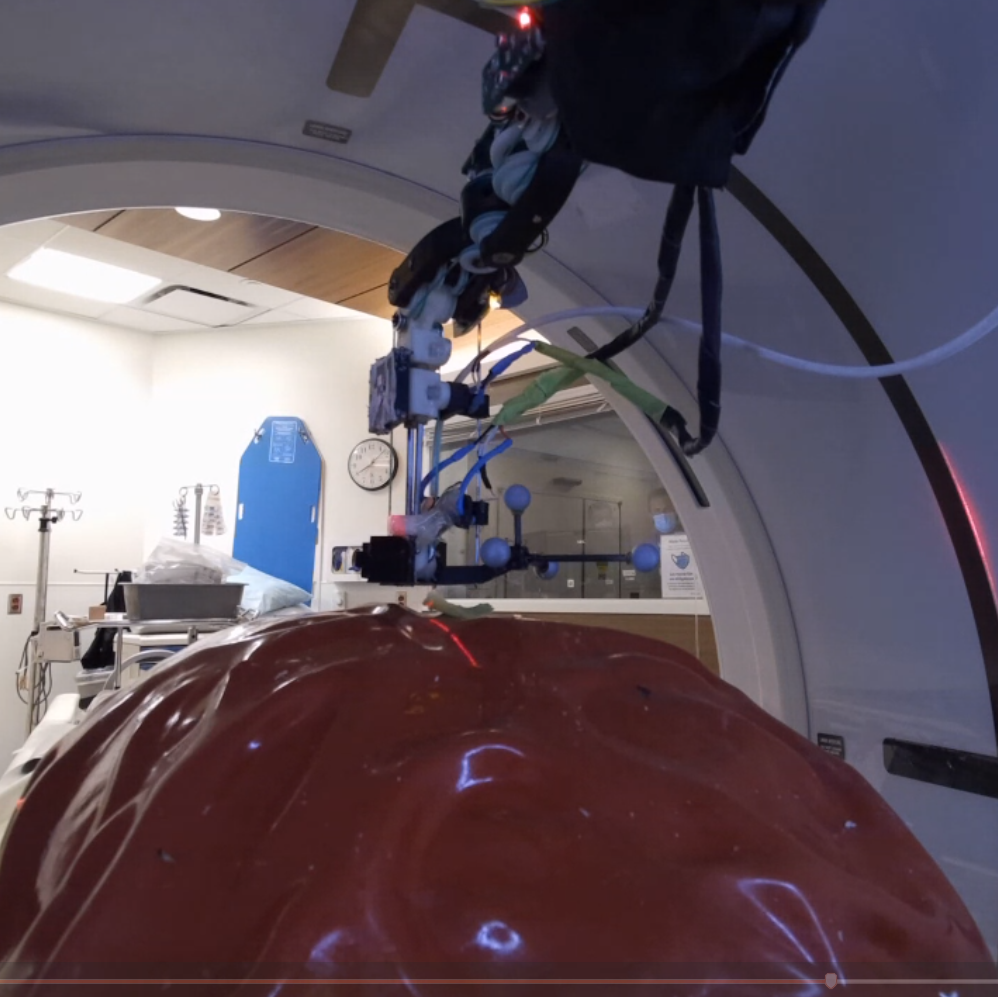}
    \includegraphics[width=0.49\columnwidth]{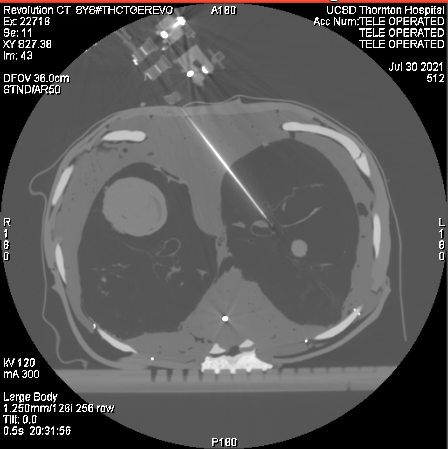}
    \caption{Left: CRANE within a CT scanner, remotely teleoperated by an experienced radiologist and technician, visible behind the glass window. Right: CT image of robot being teleoperated to reach lesion in the central right lung of the phantom. }
    \label{fig:additional_exp}
\vspace{-6mm}
\end{figure}

\section{Discussion and Conclusion}
We present a teleoperated 10-DoF, low-profile and highly
dexterous robotic needle placement platform for efficient
and accurate needle insertion across the human abdominal region. System backlash is low and through closed-loop control. End-effector error is greatly decreased providing the performance required for effective needle insertion in robotic applications. Repeatability with a trajectory tracking positional accuracy of $0.27mm$ and orientation accuracy of $0.7^{\circ}$. Minimal shadowing and artifacts are visible in the CT image. A novel clutching mechanism is included which enables long-needle insertions in an easy-to-manufacture and sterilizable assembly. In future work, we will explore collision-free path planning within the CT scanner, metrics for ranking inverse kinematics solutions at target needle-insertion vectors, and null-space control to better use the robot's dexterity to avoid collision and joint limits.



\balance
\bibliographystyle{IEEEtran}
\bibliography{references}

\begin{thebibliography}{10}
\providecommand{\url}[1]{#1}
\csname url@samestyle\endcsname
\providecommand{\newblock}{\relax}
\providecommand{\bibinfo}[2]{#2}
\providecommand{\BIBentrySTDinterwordspacing}{\spaceskip=0pt\relax}
\providecommand{\BIBentryALTinterwordstretchfactor}{4}
\providecommand{\BIBentryALTinterwordspacing}{\spaceskip=\fontdimen2\font plus
\BIBentryALTinterwordstretchfactor\fontdimen3\font minus
  \fontdimen4\font\relax}
\providecommand{\BIBforeignlanguage}[2]{{%
\expandafter\ifx\csname l@#1\endcsname\relax
\typeout{** WARNING: IEEEtran.bst: No hyphenation pattern has been}%
\typeout{** loaded for the language `#1'. Using the pattern for}%
\typeout{** the default language instead.}%
\else
\language=\csname l@#1\endcsname
\fi
#2}}
\providecommand{\BIBdecl}{\relax}
\BIBdecl

\bibitem{Tsai2009}
I.~C. Tsai, W.~L. Tsai, M.~C. Chen, G.~C. Chang, W.~S. Tzeng, S.~W. Chan, and
  C.~C.~C. Chen, ``{CT-guided core biopsy of lung lesions: A primer},''
  \emph{American Journal of Roentgenology}, vol. 193, no.~5, pp. 1228--1235, 11
  2009.

\bibitem{ctguided_retrospective}
\BIBentryALTinterwordspacing
A.~ROMAN, P.~ACHIMAS-CADARIU, B.~FETICA, V.~GATA, and A.~SEICEAN, ``{CT-guided
  procedures: an initial experience},'' \emph{Clujul Medical}, vol.~91, no.~4,
  p. 427, 10 2018. [Online]. Available: \url{/pmc/articles/PMC6296730/
  /pmc/articles/PMC6296730/?report=abstract
  https://www.ncbi.nlm.nih.gov/pmc/articles/PMC6296730/}
\BIBentrySTDinterwordspacing

\bibitem{heerink2017complication}
W.~J. Heerink, G.~H. De~Bock, G.~J. de~Jonge, H.~J.~M. Groen, .~R.
  Vliegenthart, .~M. Oudkerk, R.~Vliegenthart, and M.~Oudkerk, ``{Complication
  rates of CT-guided transthoracic lung biopsy: meta-analysis},''
  \emph{European radiology}, vol.~27, no.~1, pp. 138--148, 1 2017.

\bibitem{Accordino2015}
M.~K. Accordino, J.~D. Wright, D.~Buono, A.~I. Neugut, and D.~L. Hershman,
  ``{Trends in use and safety of image-guided transthoracic needle biopsies in
  patients with cancer},'' \emph{Journal of Oncology Practice}, vol.~11, no.~3,
  pp. e351--e359, 1 2015.

\bibitem{wang2016ct}
Y.~Wang, F.~Jiang, X.~Tan, and P.~Tian, ``{CT-guided percutaneous transthoracic
  needle biopsy for paramediastinal and nonparamediastinal lung lesions:
  Diagnostic yield and complications in 1484 patients},'' \emph{Medicine},
  vol.~95, no.~31, 2016.

\bibitem{national2013results}
N.~L. S. T.~R. Team, ``{Results of initial low-dose computed tomographic
  screening for lung cancer},'' \emph{New England Journal of Medicine}, vol.
  368, no.~21, pp. 1980--1991, 2013.

\bibitem{Tian2017}
P.~Tian, Y.~Wang, L.~Li, Y.~Zhou, W.~Luo, and W.~Li, ``{CT-guided transthoracic
  core needle biopsy for small pulmonary lesions: Diagnostic performance and
  adequacy for molecular testing},'' \emph{Journal of Thoracic Disease},
  vol.~9, no.~2, pp. 333--343, 2017.

\bibitem{kettenbach2015robotic}
J.~Kettenbach and G.~Kronreif, ``{Robotic systems for percutaneous
  needle-guided interventions},'' pp. 45--53, 2015.

\bibitem{Walsh2008}
C.~J. Walsh, N.~C. Hanumara, A.~H. Slocum, J.~A. Shepard, and R.~Gupta, ``{A
  patient-mounted, telerobotic tool for CT-guided percutaneous
  interventions},'' \emph{Journal of Medical Devices, Transactions of the
  ASME}, vol.~2, no.~1, p. 011007, 3 2008.

\bibitem{Ghelfi2018}
\BIBentryALTinterwordspacing
J.~Ghelfi, A.~Moreau-Gaudry, N.~Hungr, C.~Fouard, B.~V{\'{e}}ron, M.~Medici,
  E.~Chipon, P.~Cinquin, and I.~Bricault, ``{Evaluation of the Needle
  Positioning Accuracy of a Light Puncture Robot Under MRI Guidance: Results of
  a Clinical Trial on Healthy Volunteers},'' \emph{CardioVascular and
  Interventional Radiology}, vol.~41, no.~9, pp. 1428--1435, 9 2018. [Online].
  Available: \url{https://doi.org/10.1007/s00270-018-2001-5}
\BIBentrySTDinterwordspacing

\bibitem{Hiraki2017}
\BIBentryALTinterwordspacing
T.~Hiraki, T.~Kamegawa, T.~Matsuno, J.~Sakurai, Y.~Kirita, R.~Matsuura,
  T.~Yamaguchi, T.~Sasaki, T.~Mitsuhashi, T.~Komaki, Y.~Masaoka, Y.~Matsui,
  H.~Fujiwara, T.~Iguchi, H.~Gobara, and S.~Kanazawa, ``{Robotically driven
  CT-guided needle insertion: Preliminary results in phantom and animal
  experiments},'' \emph{Radiology}, vol. 285, no.~2, pp. 454--461, 11 2017.
  [Online]. Available: \url{https://doi.org/10.1148/radiol.2017162856}
\BIBentrySTDinterwordspacing

\bibitem{Maurin2006}
B.~Maurin, B.~Bayle, J.~Gangloff, P.~Zanne, M.~De~Mathelin, and O.~Piccin, ``{A
  robotized positioning platform guided by computed tomography: Practical
  issues and evaluation},'' in \emph{Proceedings - IEEE International
  Conference on Robotics and Automation}, vol. 2006, 2006, pp. 251--256.

\bibitem{Yang2017}
Y.~Yang, S.~Jiang, Z.~Yang, W.~Yuan, H.~Dou, W.~Wang, D.~Zhang, and Y.~Bian,
  ``{Design and analysis of a tendon-based computed tomography-compatible robot
  with remote center of motion for lung biopsy},'' \emph{Proceedings of the
  Institution of Mechanical Engineers, Part H: Journal of Engineering in
  Medicine}, vol. 231, no.~4, pp. 286--298, 2017.

\bibitem{Wu2019}
D.~Wu, G.~Li, N.~Patel, J.~Yan, R.~Monfaredi, K.~Cleary, and I.~Iordachita,
  ``{Remotely Actuated Needle Driving Device for MRI-Guided Percutaneous
  Interventions},'' 5 2019.

\bibitem{Bricault2008}
\BIBentryALTinterwordspacing
I.~Bricault, N.~Zemiti, E.~Jouniaux, C.~Fouard, Ã.~Taillant, F.~Dorandeu, and
  P.~Cinquin, ``{Light puncture robot for CT and MRI interventions},'' in
  \emph{IEEE Engineering in Medicine and Biology Magazine}, vol.~27,
  no.~3.\hskip 1em plus 0.5em minus 0.4em\relax Institute of Electrical and
  Electronics Engineers Inc., 2008, pp. 42--50. [Online]. Available:
  \url{https://pubmed.ncbi.nlm.nih.gov/18519181/}
\BIBentrySTDinterwordspacing

\bibitem{Hungr2016}
N.~Hungr, I.~Bricault, P.~Cinquin, and C.~Fouard, ``{Design and Validation of a
  CT-and MRI-Guided Robot for Percutaneous Needle Procedures},'' \emph{IEEE
  Transactions on Robotics}, vol.~32, no.~4, pp. 973--987, 2016.

\bibitem{xact_robotics}
\BIBentryALTinterwordspacing
``{Homepage - XACT Robotics}.'' [Online]. Available:
  \url{https://xactrobotics.com/}
\BIBentrySTDinterwordspacing

\bibitem{Stoianovici1998}
D.~Stoianovici, L.~L. Whitcomb, J.~H. Anderson, R.~H. Taylor, and L.~R.
  Kavoussi, ``{A modular surgical robotic system for image guided percutaneous
  procedures},'' in \emph{Lecture Notes in Computer Science (including
  subseries Lecture Notes in Artificial Intelligence and Lecture Notes in
  Bioinformatics)}, vol. 1496.\hskip 1em plus 0.5em minus 0.4em\relax Springer
  Verlag, 1998, pp. 404--410.

\bibitem{bio-rob2}
G.~Kronreif, M.~F{\"{u}}rst, W.~Ptacek, M.~Kornfeld, and J.~Kettenbach,
  ``{Robotic System for Image Guided Therapie-B-RobII}.''

\bibitem{Schulz2013}
B.~Schulz, K.~Eichler, P.~Siebenhandl, T.~Gruber-Rouh, C.~Czerny,
  T.~Josef~Vogl, and S.~Zangos, ``{Accuracy and speed of robotic assisted
  needle interventions using a modern cone beam computed tomography
  intervention suite: a phantom study}.''

\bibitem{Stoianovici1997}
\BIBentryALTinterwordspacing
D.~Stoianovici, J.~A. Cadeddu, R.~D. Demaree, S.~A. Basile, R.~H. Taylor, L.~L.
  Whitcomb, W.~N. Sharpe, and L.~R. Kavoussi, ``{An efficient needle injection
  technique and radiological guidance method for percutaneous procedures},''
  \emph{Lecture Notes in Computer Science (including subseries Lecture Notes in
  Artificial Intelligence and Lecture Notes in Bioinformatics)}, vol. 1205, pp.
  295--298, 1997. [Online]. Available:
  \url{https://link.springer.com/chapter/10.1007/BFb0029248}
\BIBentrySTDinterwordspacing

\bibitem{Masamune2001}
\BIBentryALTinterwordspacing
K.~Masamune, G.~Fichtinger, A.~Patriciu, R.~C. Susil, R.~H. Taylor, L.~R.
  Kavoussi, J.~H. Anderson, I.~Sakuma, T.~Dohi, and D.~Stoianovici, ``{System
  for Robotically Assisted Percutaneous Procedures with Computed Tomography
  Guidance},'' \emph{https://mc.manuscriptcentral.com/tcas}, vol.~6, no.~6, pp.
  370--383, 1 2001. [Online]. Available:
  \url{https://www.tandfonline.com/doi/abs/10.3109/10929080109146306}
\BIBentrySTDinterwordspacing

\bibitem{Yang2010a}
L.~Yang, R.~Wen, J.~Qin, C.~K. Chui, K.~B. Lim, and S.~K.~Y. Chang, ``{A
  robotic system for overlapping radiofrequency ablation in large tumor
  treatment},'' \emph{IEEE/ASME Transactions on Mechatronics}, vol.~15, no.~6,
  pp. 887--897, 12 2010.

\bibitem{Tovar-Arriaga2011}
\BIBentryALTinterwordspacing
S.~Tovar-Arriaga, R.~Tita, J.~C. Pedraza-Ortega, E.~Gorrostieta, and W.~A.
  Kalender, ``{Development of a robotic FD-CT-guided navigation system for
  needle placement-preliminary accuracy tests},'' \emph{International Journal
  of Medical Robotics and Computer Assisted Surgery}, vol.~7, no.~2, pp.
  225--236, 6 2011. [Online]. Available:
  \url{https://onlinelibrary.wiley.com/doi/full/10.1002/rcs.393
  https://onlinelibrary.wiley.com/doi/abs/10.1002/rcs.393
  https://onlinelibrary.wiley.com/doi/10.1002/rcs.393}
\BIBentrySTDinterwordspacing

\bibitem{jhu_fully_active_biorob}
\BIBentryALTinterwordspacing
J.~Kettenbach, G.~Kronreif, M.~Figl, M.~F{\"{u}}rst, W.~Birkfellner, R.~Hanel,
  and H.~Bergmann, ``{Robot-assisted biopsy using ultrasound guidance: initial
  results from in vitro tests},'' \emph{European Radiology 2004 15:4}, vol.~15,
  no.~4, pp. 765--771, 9 2004. [Online]. Available:
  \url{https://link.springer.com/article/10.1007/s00330-004-2487-x}
\BIBentrySTDinterwordspacing

\bibitem{Fichtinger2002}
G.~Fichtinger, T.~L. DeWeese, A.~Patriciu, A.~Tanacs, D.~Mazilu, J.~H.
  Anderson, K.~Masamune, R.~H. Taylor, and D.~Stoianovici, ``{System for
  Robotically Assisted Prostate Biopsy and Therapy with Intraoperative CT
  Guidance},'' \emph{Academic Radiology}, vol.~9, no.~1, pp. 60--74, 1 2002.

\bibitem{korea_cool_robot}
H.~Jin Won Namkug Kim Guk Bae Kim Joon Beom Seo Hongho~Kim, ``{Validation of a
  CT-guided intervention robot for biopsy and radiofrequency ablation:
  experimental study with an abdominal phantom From the Departments of
  Radiology ( I N T E R V E N T I O N A L R A D I O LO G Y},'' 2017.

\bibitem{Moreira2017a}
P.~Moreira, G.~van~de Steeg, T.~Krabben, J.~Zandman, E.~E.~G. Hekman,
  F.~van~der Heijden, R.~Borra, and S.~Misra, ``{The MIRIAM Robot: A Novel
  Robotic System for MR-Guided Needle Insertion in the Prostate},''
  \emph{Journal of Medical Robotics Research}, vol.~02, no.~04, p. 1750006, 12
  2017.

\bibitem{Shahriari2015}
\BIBentryALTinterwordspacing
N.~Shahriari, E.~Hekman, M.~Oudkerk, and S.~Misra, ``{Design and evaluation of
  a computed tomography (CT)-compatible needle insertion device using an
  electromagnetic tracking system and CT images},'' \emph{International Journal
  of Computer Assisted Radiology and Surgery}, vol.~10, no.~11, pp. 1845--1852,
  11 2015. [Online]. Available: \url{/pmc/articles/PMC4617842/
  /pmc/articles/PMC4617842/?report=abstract
  https://www.ncbi.nlm.nih.gov/pmc/articles/PMC4617842/}
\BIBentrySTDinterwordspacing

\bibitem{Stoianovici2003}
D.~Stoianovici, K.~Cleary, A.~Patriciu, D.~Mazilu, A.~Stanimir, N.~Craciunoiu,
  V.~Watson, and L.~Kavoussi, ``{AcuBot: A Robot for Radiological
  Interventions},'' \emph{IEEE Transactions on Robotics and Automation},
  vol.~19, no.~5, pp. 927--930, 2003.

\bibitem{number25}
A.~Melzer, B.~Gutmann, T.~Remmele, R.~Wolf, A.~Lukoscheck, M.~Bock,
  H.~Bardenheuer, and H.~Fischer, ``{INNOMOTION for Percutaneous Image-Guided
  Interventions},'' \emph{IEEE Engineering in Medicine and Biology Magazine},
  vol.~27, no.~3, pp. 66--73, 5 2008.

\bibitem{frishman2021}
S.~Frishman, R.~D. Ings, V.~Sheth, B.~L. Daniel, and M.~R. Cutkosky,
  ``{Extending Reach Inside the MRI Bore: A 7-DOF, Low-Friction, Hydrostatic
  Teleoperator},'' \emph{IEEE Transactions on Medical Robotics and Bionics},
  vol.~3, no.~3, pp. 701--713, 8 2021.

\bibitem{kapoor2008}
\BIBentryALTinterwordspacing
S.~Shah, A.~Kapoor, J.~Ding, P.~Guion, D.~Petrisor, J.~Karanian, W.~F.
  Pritchard, D.~Stoianovici, B.~J. Wood, and K.~Cleary, ``{Robotically assisted
  needle driver: evaluation of safety release, force profiles, and needle spin
  in a swine abdominal model},'' \emph{International Journal of Computer
  Assisted Radiology and Surgery 2008 3:1}, vol.~3, no.~1, pp. 173--179, 5
  2008. [Online]. Available:
  \url{https://link.springer.com/article/10.1007/s11548-008-0164-2}
\BIBentrySTDinterwordspacing

\bibitem{Schreiber2019}
D.~D.~A. Schreiber, D.~D.~B. Shak, A.~M.~A. Norbash, and M.~C. Yip, ``{An
  Open-Source 7-Axis, Robotic Platform to Enable Dexterous Procedures within CT
  Scanners},'' in \emph{IEEE International Conference on Intelligent Robots and
  Systems}.\hskip 1em plus 0.5em minus 0.4em\relax Institute of Electrical and
  Electronics Engineers Inc., 11 2019, pp. 386--393.

\bibitem{solomon2002robotically}
\BIBentryALTinterwordspacing
S.~B. Solomon, A.~Patriciu, M.~E. Bohlman, L.~R. Kavoussi, and D.~Stoianovici,
  ``{Robotically driven interventions: A method of using CT fluoroscopy without
  radiation exposure to the physician},'' \emph{Radiology}, vol. 225, no.~1,
  pp. 277--282, 10 2002. [Online]. Available:
  \url{/pmc/articles/PMC3107539/?report=abstract
  https://www.ncbi.nlm.nih.gov/pmc/articles/PMC3107539/}
\BIBentrySTDinterwordspacing

\bibitem{Frishman2020}
S.~Frishman, A.~Kight, I.~Pirozzi, M.~C. Coffey, B.~L. Daniel, and M.~R.
  Cutkosky, ``{Enabling In-Bore MRI-Guided Biopsies with Force Feedback},''
  \emph{IEEE Transactions on Haptics}, vol.~13, no.~1, pp. 159--166, 1 2020.

\bibitem{Poniatowski2016}
L.~H. Poniatowski, S.~S. Somani, D.~Veneziano, S.~McAdams, and R.~M. Sweet,
  ``{Characterizing and Simulating Needle Insertion Forces for Percutaneous
  Renal Access},'' \emph{Journal of Endourology}, vol.~30, no.~10, pp.
  1049--1055, 10 2016.

\bibitem{Walsch_thesis_2010}
\BIBentryALTinterwordspacing
C.~J. Walsh, ``{Image-guided robots for dot-matrix tumor ablation},'' \emph{A.
  Mechanical and Manufacturing Engineering}, 2010. [Online]. Available:
  \url{https://dspace.mit.edu/handle/1721.1/61613}
\BIBentrySTDinterwordspacing

\bibitem{Scott1992}
W.~W. Scott and J.~E. Kuhlman, ``{Phantom for use in lung biopsy training},''
  \emph{Radiology}, vol. 184, no.~1, pp. 286--287, 1992.

\end{thebibliography}

\end{document}